\documentclass{article}
\usepackage[utf8]{inputenc}
\usepackage{main}
\usepackage{microtype}
\usepackage{graphicx}
\usepackage{times}
\usepackage{amsmath}
\usepackage{amssymb}
\usepackage{enumitem}
\usepackage{booktabs}
\usepackage{xcolor}     
\usepackage{natbib}
\usepackage{wrapfig}
\usepackage{multirow}
\usepackage[table]{xcolor}
\definecolor{mydarkblue}{rgb}{0,0.08,0.45}
\usepackage[colorlinks=true,linkcolor=mydarkblue,citecolor=mydarkblue,filecolor=mydarkblue,urlcolor=mydarkblue]{hyperref}
\usepackage{fancyhdr}

\usepackage[most]{tcolorbox}
\tcbuselibrary{breakable}

\definecolor{findbg}{RGB}{245,250,244}
\definecolor{findframe}{RGB}{78,121,72}
\definecolor{insightbg}{RGB}{241,246,253}
\definecolor{insightframe}{RGB}{67,104,163}
\definecolor{impbg}{RGB}{255,247,237}
\definecolor{impframe}{RGB}{181,114,37}

\newtcolorbox[auto counter]{finding}[1][] {
  colback=findbg,
  colframe=findframe,
  fonttitle=\bfseries,
  title=Finding~\thetcbcounter,
  enhanced,
  breakable,
  boxrule=0.5pt,
  left=1mm,right=1mm,top=1mm,bottom=1mm,
  #1
}

\newtcolorbox[auto counter]{insight}[1][] {
  colback=insightbg,
  colframe=insightframe,
  fonttitle=\bfseries,
  title=Insight~\thetcbcounter,
  enhanced,
  breakable,
  boxrule=0.5pt,
  left=1mm,right=1mm,top=1mm,bottom=1mm,
  #1
}

\newtcolorbox[auto counter]{implication}[1][] {
  colback=impbg,
  colframe=impframe,
  fonttitle=\bfseries,
  title=Implication~\thetcbcounter,
  enhanced,
  breakable,
  boxrule=0.5pt,
  left=1mm,right=1mm,top=1mm,bottom=1mm,
  #1
}

\newenvironment{itemize*}%
 {\leftmargini=10pt\begin{itemize}%
  \setlength{\itemsep}{0pt}%
  \setlength{\parskip}{0pt}%
  }%
 {\end{itemize}}
\newenvironment{enumerate*}%
 {\begin{enumerate}%
  \setlength{\itemsep}{0pt}%
  \setlength{\parskip}{0pt}}%
 {\end{enumerate}}

\begin{document}

\title{Rethinking RL for LLM Reasoning:\\ It’s Sparse Policy Selection, Not Capability Learning} 
% \title{Unveiling the Dark Side of RL-Incentivized Reasoning: A Controlled Study of Pre-, Mid-, and Post-Training Interplay}
% \title{Demystify the Dark Side of RL-Incentivized Reasoning: A Controlled Study of Pre-, Mid-, and Post-Training Interplay}
% \title{Demystifying LM Reasoning: A Controlled Study of Pre-Training, Mid-Training and RL Interplay} 
% \title{Unveiling the Art of RL-Incentivized Reasoning: A Controlled Study of Pre-, Mid-, and Post-Training Interplay}
% \title{Demystify the Art of RL-Incentivized Reasoning: A Controlled Study of Pre-, Mid-, and Post-Training Interplay}
% \title{On the Interplay of Pre-Training, Mid-Training, and RL \\in Language Model Reasoning
% }
% \title{When Does RL Incentivize Reasoning Beyond the Base Model? Interplay of Pre-, Mid-, and Post-training in LM Reasoning}
% \title{When Does RL Incentivize Reasoning Beyond the Base Model? A Controlled Study of Pre-, Mid-, and Post-training Interplay}

\author{
\textbf{Ömer Faruk Akgül}$^{1}$ \quad
\textbf{Rajgopal Kannan}$^{2}$ \quad
\textbf{Willie Neiswanger}$^{1}$ \quad
\textbf{Viktor Prasanna}$^{1}$ \\[5pt]
$^{1}$University of Southern California \quad
$^{2}$DEVCOM ARL \\[4pt]
\href{https://github.com/farukakgul/ReasonMaxxer}{\includegraphics[height=0.4cm]{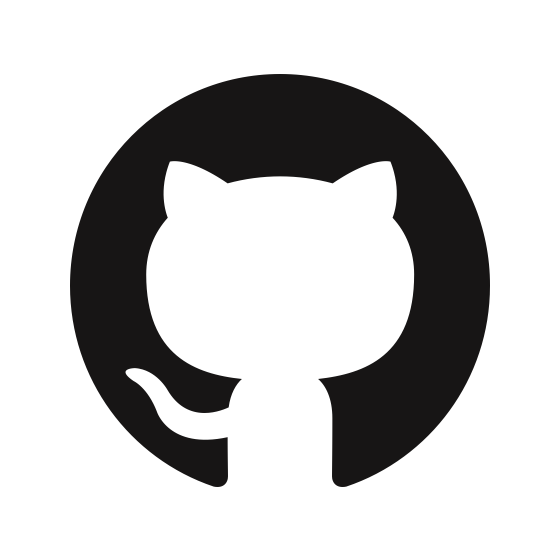} \textbf{\textsc{ReasonMaxxer}: Embarrassingly Cheap Post-Training}} \\
}

\maketitle
\thispagestyle{fancy}
\fancyhead{}
\lhead{\includegraphics[height=0.85cm]{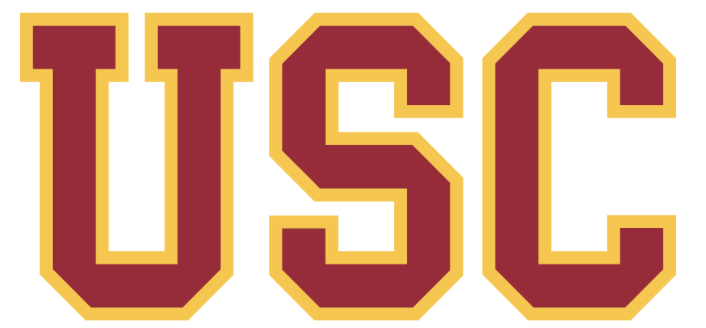}}
\rhead{{\textit{May 7, 2026}}}

\renewcommand{\headrulewidth}{0pt}
\setlength{\headheight}{14pt}
\addtolength{\topmargin}{3pt}
\setlength{\headsep}{3mm}

\begin{abstract} Reinforcement learning has become the standard for improving reasoning in large language models, yet evidence increasingly suggests that RL does not teach new strategies; it redistributes probability mass over solutions the base model already contains.  In this work we ask: if RL merely steers the model toward paths it already knows, is the RL optimization loop itself necessary?  Through token‑level analysis across multiple model families and RL algorithms, we find that RL's beneficial footprint is a sparse, predictable correction concentrated at high‑entropy \emph{decision points} where the model is uncertain which branch to take.  Only 1--3\% of token positions are affected, the promoted token always lies within the base model's top‑5 alternatives, and targeted corrections at those few positions causally recover a large fraction of RL's accuracy gain, while random corrections fail.  The base model's own entropy identifies these positions without any RL‑trained model, and the entire correction is low‑dimensional, representable in a tiny fraction of model parameters.  These findings reframe reasoning improvement as sparse \emph{policy selection}, not capability acquisition.  We translate this insight into \textsc{ReasonMaxxer}, a minimal RL‑free method that applies contrastive loss only at entropy‑gated decision points, using a few hundred base‑model rollouts and no online generation.  Across three model families, six scales, and six math reasoning benchmarks, \textsc{ReasonMaxxer} matches or exceeds full RL performance while requiring only tens of problems and minutes of single‑GPU training, a \textit{reduction in training cost of roughly three orders of magnitude}. \end{abstract}

\section{Introduction}
\label{sec:intro}

\begin{figure}[ht]
\centering
\includegraphics[width=0.60\columnwidth]{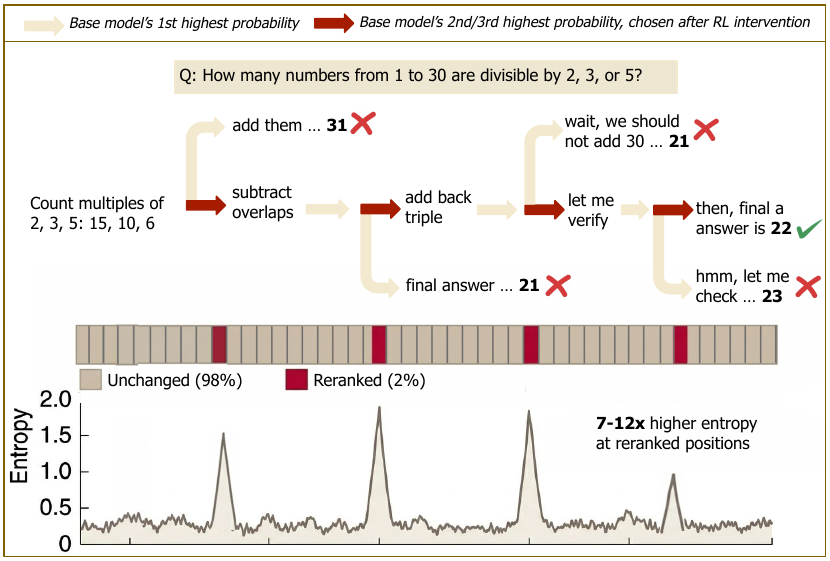}
\caption{\textbf{RL edits are rare, conservative, and concentrated at decision points.}
\textbf{(a)} The RL model's chosen token is on average rank~2 among the base model's
top alternatives, meaning it almost never invents a new token but instead promotes
one the base model was already considering.
\textbf{(b)} Only 1--4\% of token positions are reranked by RL, yet those positions
have higher base‑model entropy than unchanged positions.  The sparse
edits thus land exactly at high‑uncertainty \emph{decision points} where the model
is unsure which reasoning branch to take.}
\label{fig:sec3_overview}
\end{figure}

Reinforcement learning with verifiable rewards (RLVR) has become the dominant paradigm for improving reasoning in large language models~\citep{guo2025deepseek, shao2024deepseekmath, zeng2025simplerl}. Systems such as DeepSeek-R1~\citep{guo2025deepseek}, OpenAI o1~\citep{jaech2024openai}, and Qwen3~\citep{qwen2025qwen3} demonstrate substantial gains from this pipeline, and the field has broadly adopted RL, typically GRPO~\citep{shao2024deepseekmath} or PPO~\citep{schulman2017ppo}, as the standard post-training method for mathematical and code reasoning. The implicit assumption underlying this paradigm is that RL, similar to how it discovers novel strategies in games~\citep{silver2017mastering}, enables LLMs to acquire genuinely new reasoning patterns through reward-driven exploration.  
A growing body of evidence challenges this assumption. \citet{yue2025does} show that while RL improves pass@$1$, base models achieve higher pass@$k$ at large $k$: the base model's sampling distribution already contains correct solutions that RL merely promotes. \citet{davis2025objective} prove that popular RL algorithms with binary rewards all reduce to stochastic gradient ascent on monotone transforms of the probability of a correct answer, and that such optimization is only profitable when the base model already succeeds non-trivially. \citet{zhang2025interplay} confirm this through controlled experiments: RL produces genuine gains only at the model's edge of competence, on problems that are difficult but not yet out of reach. At the token level, \citet{wang2025beyond} identify that RL's improvements concentrate at high-entropy ``forking tokens'' where the model is uncertain which reasoning path to follow, and show that restricting gradient updates to these tokens matches training on all tokens. From a structural angle, \citet{park2025thinking} find that RL operates through a small number of emergent attention heads. Collectively, these findings converge on an emerging picture: \emph{RL primarily steers the model toward committing to solution paths that the base model already contains, rather than inventing genuinely new reasoning strategies.}  

Despite this growing understanding, a critical gap remains.  The works that identify this structure still operate inside the RL framework: \citet{wang2025beyond} make RL more efficient rather than eliminating it, \citet{yue2025does} call for improved RL paradigms, and \citet{karan2025sampling} offer only inference‑time alternatives.  The natural next question is \textit{whether we can precisely characterize RL's token‑level effect and, if that characterization is simple enough, whether the RL optimization loop itself is necessary}.

In this paper, we answer that question through a systematic token-level analysis across multiple model families and RL algorithms. We find that RL's behavioral footprint is strikingly simple: it modifies only 1--3\% of token positions, does not introduce tokens outside the base model's top-5 candidates, and concentrates edits at high-entropy \emph{decision points} where the model is uncertain which reasoning branch to take. Using oracle intervention with random controls, we establish that the specific token chosen at these positions matters causally, recovering a large share of RL's gain, while random corrections fail. Crucially, these decision points can be located without any RL-trained model: the base model's own token entropy, which peaks at the positions RL edits, provides a strong proxy for where intervention is useful. We further show that the full correction is low-dimensional, representable in a tiny fraction of model parameters. Together, these findings reframe reasoning improvement as a \textbf{sparse \emph{policy selection} problem: committing to the right branch at a handful of uncertainty points}, rather than acquiring new capabilities through expensive exploration.

To test this reframing directly, we construct \textsc{ReasonMaxxer}, a minimal RL-free method that exploits the identified structure. \textsc{ReasonMaxxer} generates a small set of rollouts from the base model, uses entropy gating to locate decision points, and applies an advantage-weighted contrastive loss exclusively at those positions, while anchoring all other tokens to the base distribution. The method requires no RL, no online generation, and no large-scale compute: it maximizes reasoning performance with a shoestring budget. Across three model families and multiple scales, \textsc{ReasonMaxxer} matches or exceeds the performance of models trained with full RL, yet uses only tens of problems, hundreds of rollouts, and minutes of single-GPU training, \textbf{reducing training cost by roughly three orders of magnitude}. That so simple a method suffices challenges the prevailing assumption that heavy RL infrastructure is necessary for reasoning improvement. 

Our contributions are as follows: \begin{itemize}[leftmargin=1.5em, itemsep=3pt, topsep=3pt]   \item \textbf{Mechanistic characterization of RL for reasoning.}   Through token‑level analysis across multiple model families and RL algorithms,   we show that RL's beneficial effect is a sparse, entropy‑localized reranking of   tokens the base model already favors, and we establish causality through oracle   intervention with random controls.   \item \textbf{An RL‑free method that matches full RL.}   We introduce \textsc{ReasonMaxxer}, which applies contrastive fine‑tuning only   at entropy‑gated decision points using the base model's own rollouts.  It   matches or exceeds RL‑trained models on math reasoning benchmarks while using   orders‑of‑magnitude less compute and data.   \item \textbf{Evidence that heavy RL is not a prerequisite.} By showing that a lightweight method can replicate RL's reasoning improvement, we demonstrate that the problem RL solves in this domain is sparse policy   selection, not capability acquisition.  This suggests that the \textit{community's default investment in full RL pipelines for outcome‑based reasoning may be excessive relative to the problem's complexity}. \end{itemize}
\section{Background and Experimental Setup} 
\label{sec:background}  

\subsection{Reinforcement Learning with Verifiable Rewards} \label{sec:rlvr}  We briefly review the RL algorithms used by the baseline models in our study. Given a prompt $q$ with ground-truth answer $a$, RLVR generates $G$ rollouts $\{o^i\}_{i=1}^G$ from the current policy $\pi_\theta$ and assigns each a binary reward $R^i = \mathbf{1}[\texttt{match}(o^i, a)]$.  The dominant algorithm among the baselines we evaluate is Group Relative Policy Optimization (GRPO)~\citep{shao2024deepseekmath}, which computes per-rollout advantages via group normalization: \begin{equation} \hat{A}^i = \frac{R^i - \text{mean}(\{R^j\}_{j=1}^G)}{\text{std}(\{R^j\}_{j=1}^G)}, \label{eq:grpo_adv} \end{equation} and updates the policy by maximizing a clipped surrogate objective applied uniformly across all token positions. This uniform application is a key point of contrast with our approach: GRPO distributes gradient across every token in every rollout, despite the evidence (presented in \S\ref{sec:post}) that only a small fraction of positions carry the useful signal.  Several baselines use alternative algorithms that share the same core structure. Open-Reasoner-Zero~\citep{hu2025openreasonerzero} employs Proximal Policy Optimization (PPO)~\citep{schulman2017ppo} with GAE, while other recent work explores REINFORCE-style variants such as RLOO~\citep{ahmadian2024back}. All of these methods optimize the same underlying objective: increasing the probability of tokens that lead to correct answers, with the primary differences lying in advantage estimation and regularization strategies. Our mechanistic analysis in \S\ref{sec:post} studies models trained with GRPO, PPO, and RLOO, and finds the same sparse-correction pattern across all three.

\subsection{Token-Level Entropy and Decision Points} \label{sec:entropy_def}  For an autoregressive language model $\pi_\theta$, the token-level generation entropy at position $t$ is defined as \begin{equation} H_t = -\sum_{v \in \mathcal{V}} \pi_\theta(v \mid q, o_{<t}) \log \pi_\theta(v \mid q, o_{<t}), \label{eq:entropy} \end{equation} where $\mathcal{V}$ is the vocabulary and $o_{<t}$ denotes the tokens generated so far. Positions with high $H_t$ correspond to points where the model distributes probability mass across multiple plausible continuations rather than committing to a single token. Recent work has identified these high-entropy positions as functionally significant: \citet{wang2025beyond} show that they act as ``forks'' steering the model toward different reasoning pathways, and \citet{agarwal2025entropy} demonstrate that minimizing entropy without labeled data can improve reasoning performance. We refer to positions where $H_t$ exceeds a threshold $\tau$ as \emph{decision points}, the subset of the generation where the model's commitment to a reasoning path is genuinely uncertain.

\subsection{Models and Baselines}
\label{sec:models}

The RL algorithms used by the baselines were introduced in Section~\ref{sec:rlvr} (GRPO, PPO, and their variants).  Table~\ref{tab:models_baselines} summarises the model families and the specific publicly available RL‑trained checkpoints we used in our experiments across the paper.  All baselines are trained with verifiable outcome rewards on mathematical reasoning problems.

\definecolor{creamlight}{RGB}{255,250,240}

\begin{table}[ht]
\centering
\caption{\textbf{Model families and RL baselines.}  Each baseline is a publicly
available checkpoint trained with verifiable outcome rewards on mathematical
reasoning problems.  The ``Algorithm \& note'' column identifies the RL variant
and any notable training detail.}\label{tab:models_baselines}
\small 
\setlength{\tabcolsep}{4pt}
\rowcolors{2}{}{}
\begin{tabular}{@{}l l l l@{}}
\toprule
\textbf{Family} & \textbf{Base models} & \textbf{RL baseline} & \textbf{Algorithm \& note} \\
\midrule
Qwen2.5 & 1.5B, 7B, Math‑7B, 32B & SimpleRL‑Zoo~\citep{zeng2025simplerl} & GRPO \\
        & 1.5B, 7B, 32B & Open‑Reasoner‑Zero~\citep{hu2025openreasonerzero} & PPO \\
        & Math‑7B & Eurus‑2‑7B‑PRIME~\citep{cui2025prime} & RLOO \\
\cmidrule{1-4}
Qwen3   & 0.6B & GRPO (raw base)~\citep{qwen2025qwen3} & GRPO \\
        & 4B   & General‑Reasoner~\citep{ma2025generalreasoner} & GRPO + verifier, multi‑domain \\
\cmidrule{1-4}
DeepSeek & R1‑Distill‑Qwen‑1.5B & DeepScaleR~\citep{luo2025deepscaler} & GRPO, context scaling \\
         &                       & STILL‑3~\citep{min2024still}   & PPO \\
         &                       & Open‑RS3~\citep{dang2025openrs3} & GRPO, tight compute \\
\cmidrule{1-4}
Mistral  & 7B v0.1 & SimpleRL‑Zoo~\citep{zeng2025simplerl} & GRPO \\
\bottomrule
\end{tabular}
\end{table}
\section{What RL Actually Changes: Sparse Corrections at Decision Points}
\label{sec:post}

Recent work suggests that RL for reasoning primarily steers the model toward
solutions it already knows rather than inventing new strategies~\citep{yue2025does,
davis2025objective, zhang2025interplay}.  To understand \emph{what} this steering
looks like at the token level, we compare the outputs of a base model and its
RL‑trained counterpart on the same set of prompts.

Our investigation addresses three questions:

\begin{enumerate}[itemsep=1pt, leftmargin=2.5em]
  \item \textbf{How often, and at what kind of positions, does the RL model disagree
        with the base model?} (\S\ref{sec:sparse})
  \item \textbf{Do these token‑level disagreements \emph{cause} the observed accuracy
        gain?} (\S\ref{sec:oracle})
  \item \textbf{Can we locate the critical positions without access to the RL model,
        using only signals from the base model?} (\S\ref{sec:entropy_proxy})
\end{enumerate}
We focus on the four base/RL‑tuned pairs introduced in \S\ref{sec:models} and
evaluate on MATH‑500 with deterministic decoding ($T{=}0$).

\subsection{Disagreement Is Rare, Conservative, and Concentrated at Decision Points}
\label{sec:sparse}

For each prompt, we generate a response from the base model and, at every token
position, we record which token the RL‑tuned \emph{teacher} model would have
preferred given the identical prefix.  Positions are then classified as follows:
\begin{equation}
\begin{aligned}
\textsc{Unshifted}:\;& \arg\max \pi_{\text{base}} = \arg\max \pi_{\text{teacher}}, \\
\textsc{Reranked}:\;& \arg\max \pi_{\text{teacher}} \neq \arg\max \pi_{\text{base}},\;
\arg\max \pi_{\text{teacher}} \in \mathrm{Top}\hbox{-}5(\pi_{\text{base}}), \\
\textsc{Shifted}:\;& \arg\max \pi_{\text{teacher}} \notin \mathrm{Top}\hbox{-}5(\pi_{\text{base}}).
\end{aligned}
\end{equation}
In words, \textsc{reranked} means the teacher promotes a token that was already
among the base model's top‑5 candidates, whereas \textsc{shifted} would indicate a
genuinely new preference.

\begin{table}[ht]
\centering
\caption{\textbf{Token-level divergence is minimal and localized.} Reranked positions are rare. Almost all promoted tokens remain within the base model’s top-5, the mean rank of the teacher’s token is around 2, and reranked positions have much higher base-model entropy than unchanged ones.}
\label{tab:sec3_taxonomy}
\setlength{\tabcolsep}{3.8pt}
\begin{tabular}{@{}lcccc@{}}
\toprule
\textbf{Pair} & \textbf{Reranked} & \textbf{Shifted} & \textbf{Entropy ratio} & \textbf{Mean rank} \\
\midrule
Qwen2.5-1.5B $\rightarrow$ \textcolor[HTML]{990000}{GRPO}
  & 2.09\% & 0.02\% & 7.58$\times$ & 2.30 \\
Qwen2.5-7B $\rightarrow$ \textcolor[HTML]{990000}{GRPO}
  & 1.03\% & 0.01\% & 8.27$\times$ & 2.14 \\
Qwen2.5-7B $\rightarrow$ \textcolor[HTML]{990000}{PPO}
  & 3.96\% & 0.12\% & 9.41$\times$ & 2.39 \\
Qwen3-4B $\rightarrow$ \textcolor[HTML]{990000}{GRPO}
  & 2.29\% & 0.02\% & 12.63$\times$ & 2.25 \\
\bottomrule
\end{tabular}
\end{table}

The results, summarized in Fig.~\ref{fig:sec3_overview} and
Table~\ref{tab:sec3_taxonomy}, paint a clear picture.  Only 1.0--4.1\% of all
token positions are reranked, and we observe zero shifted positions in any pair.
The teacher's preferred token is, on average, the second most likely token under
the base model (mean rank 2.14--2.39).  Moreover, the reranked positions have
5--12$\times$ higher base‑model entropy than unchanged positions.  Thus, RL's
edits are not only extremely sparse; they are also highly predictable: they occur
exactly at high‑entropy \emph{decision points} where the model is uncertain which
reasoning branch to follow (cf.\ \S\ref{sec:entropy_def}). \textit{RL does not introduce novel tokens; it consistently elevates one of the base model's top alternatives at moments of uncertainty.} This explains why prior work observed low perplexity between RL‑trained and base models~\citet{yue2025does}: the
promoted token was already a plausible candidate.

\subsection{Correcting Only the Disagreements Recovers RL Performance}
\label{sec:oracle}

Having established \emph{where} the two models differ, we now ask whether these
differences are causally responsible for the RL model's higher accuracy.  We design
an oracle intervention: during deterministic generation from the base model, at
every position where the teacher disagrees (i.e., the reranked positions from
Table~\ref{tab:sec3_taxonomy}), we replace the base token with the teacher's
preferred token and continue generating from the corrected prefix.  As a control,
we instead insert a randomly chosen alternative from the base model's top‑20
(\emph{random substitution}).

\begin{figure}[ht]
\centering
\includegraphics[width=0.77\columnwidth]{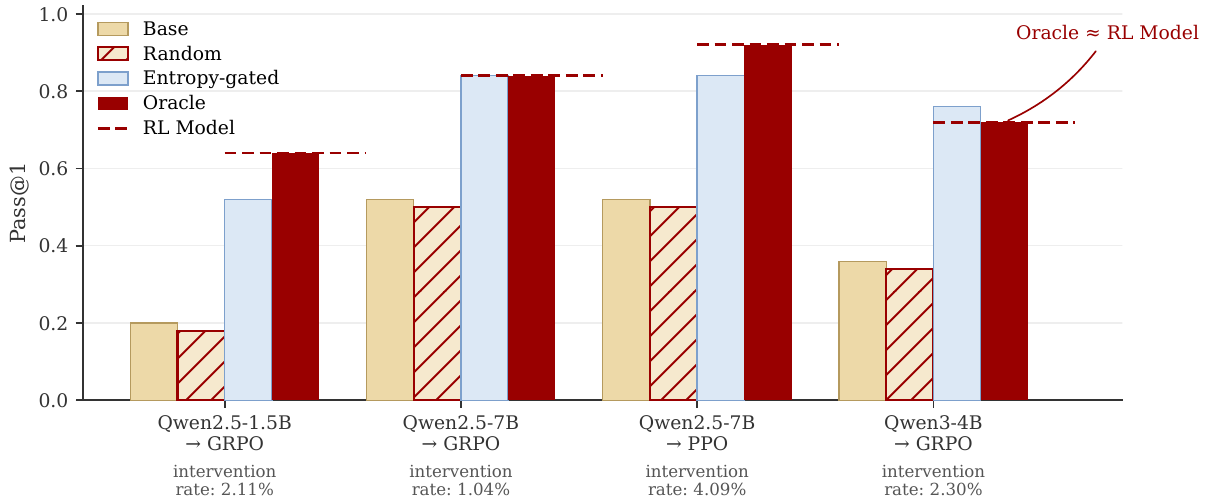}

\caption{\textbf{Oracle correction recovers RL performance exactly, and entropy-gating largely matches it.} The cream and dashed cream bars show base and random substitution; the red oracle bar matches the dashed RL model line precisely, using only 1–4\% of tokens. The cardinal red entropy-gated condition achieves comparable or identical accuracy with a similar budget, using only base-model entropy to choose where to intervene.}
\label{fig:sec3_oracle}

\end{figure}

Figure~\ref{fig:sec3_oracle} shows the outcome.  The oracle intervention
reproduces the teacher's pass@1 \emph{exactly} on every pair, while the random
substitution baseline performs no better than the base model (often worse).  The
fraction of tokens touched by the oracle equals the rerank percentages from
Table~\ref{tab:sec3_taxonomy} (1.0--4.1\%).  Hence, the RL model's entire
accuracy advantage can be attributed to a tiny set of precise token choices at
decision points. In short, \textit{a handful of token corrections can redirect the full reasoning trajectory}; RL's benefit is not a diffuse effect but is concentrated at a few branch points where the choice of continuation determines the solution path.

\subsection{Entropy Alone Identifies the Critical Positions}
\label{sec:entropy_proxy}

The oracle experiment relies on the teacher to both locate and correct the
important tokens.  For a practical RL‑free method, we need to locate these
positions without the teacher.  The strong correlation observed in \S\ref{sec:sparse}
suggests that base‑model entropy might serve this role.  We therefore test an
\emph{entropy‑gated} intervention: we replace the base token with the teacher's
preferred token at every position where the base‑model entropy exceeds a threshold
$\tau$, \emph{without} using any information about the teacher's preferences.  This
probe tells us how well entropy alone can substitute for the teacher's knowledge of
\emph{where} to intervene.

The blue bars in Fig.~\ref{fig:sec3_oracle} show the
performance of this entropy‑gated correction.  With only an entropy threshold
($\tau = $ 1.2), the intervention matches the
teacher exactly on the 7B GRPO pair, closely approaches it on the PPO pair, and
substantially improves over the base model on the other pairs, while touching only
1.2--8.3\% of tokens.  Entropy therefore acts as an effective, fully teacher‑free
proxy for the decision points that RL would correct. Thus, \textit{the \emph{where} of RL's correction is predictable from the base model's entropy alone}; the remaining challenge is to learn \emph{which} token to substitute at those positions, a problem we solve with \textsc{ReasonMaxxer} (\S\ref{sec:method}).
\section{The Correction Is Low-Dimensional}
\label{sec:compress}

Section~\ref{sec:post} showed that RL's beneficial effect is sparse in token space and
predictable from the base model's entropy.  A natural next question is whether the
correction is also simple in parameter space.  If replicating the RL model's behavior
at decision points required high‑dimensional parameter changes, the observed token‑level
sparsity might be an emergent property of a complex distributed computation, and the full
RL optimization loop might still be necessary.  Several studies have noted that such
large‑scale RL can produce representations that look low‑dimensional only after the
fact~\citep{park2025thinking}.  To test whether RL's \emph{correction} is inherently
low‑dimensional, we measure how much adapter capacity is needed to capture it.

\subsection{Distilling RL into a Low‑Rank Adapter}
\label{sec:compress_setup}

Our diagnostic is a \textbf{KL‑LoRA} distillation: we attach a LoRA adapter~\citep{hu2021lora}
to the base model and train only the adapter parameters to minimise the token‑level
Kullback–Leibler divergence between the adapter‑augmented model and the RL‑trained
teacher:
\begin{equation}
\mathcal{L}_{\text{distill}} = \sum_t
\text{KL}\!\left(\pi_{\text{teacher}}(\cdot \mid x_{<t})
\;\|\; \pi_{\text{base}+\Delta\theta}(\cdot \mid x_{<t})\right).
\label{eq:sec4_kl}
\end{equation}
We cache the teacher's top‑$k$ logits on a set of rollouts generated by the teacher
itself.  The adapters are trained on only \textbf{100 randomly chosen problems}.  If a
tiny adapter can absorb RL's full distributional change from such a small number of
problems, then that change must be fundamentally low‑dimensional.

\subsection{A Small Adapter Captures RL's Full Correction}
\label{sec:compress_results}

\begin{figure}[ht]
\centering
\includegraphics[width=0.8\columnwidth]{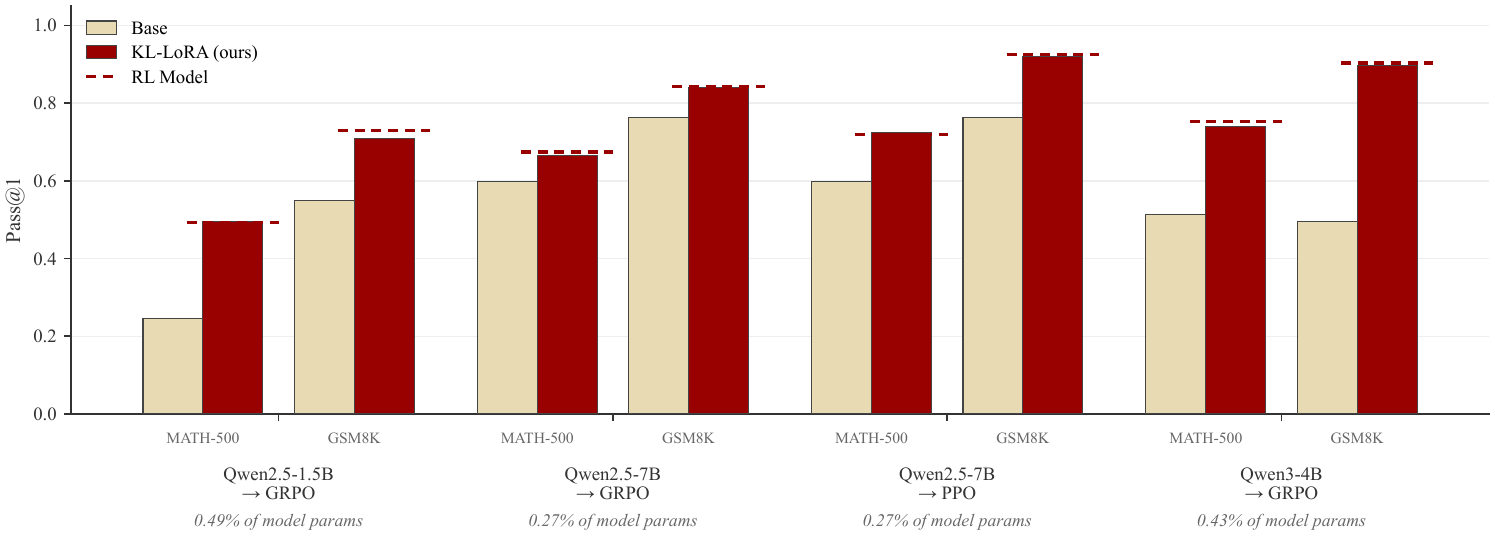}
\caption{\textbf{RL's correction is low-dimensional.} A LoRA adapter ($W_{QKVO}$, rank 32) distilled from the RL teacher via KL divergence on just 100 randomly chosen problems reproduces the teacher’s accuracy on MATH-500 and GSM8K across all four model pairs. The cream bars (base model) are far below RL teacher and the cardinal red bars (KL-LoRA) matches RL model's performance. The percentage below each group indicates the fraction of model parameters used by the adapter.}
\label{fig:sec4_main}
\end{figure}

Figure~\ref{fig:sec4_main} presents the results for the four base/RL pairs studied in
\S\ref{sec:post}.  On both MATH‑500 and GSM8K, a LoRA adapter with rank~32 applied to
all attention projections (QKVO) matches the RL teacher's accuracy, while modifying
only 0.27--0.49\% of the base model's parameters.

The adapter sizes above each group (0.3\% to 0.5\%) make the low‑dimensional nature of RL's correction immediately visible.  The design is frugal by intent: using only 100 randomly chosen problems, the adapter sees just enough examples of the model's behaviour at critical decision points to capture RL's policy steering.  This reinforces the insight from \S\ref{sec:post} that RL's signal is concentrated in a few high‑entropy locations; a small, targeted dataset suffices because the base model already possesses the necessary vocabulary and reasoning patterns.\footnote{Further compression is possible: a rank‑8 output‑projection adapter matches the full $W_{QKVO}$ adapter within a few points on MATH‑500 (Appendix~\ref{app:kl_details}), indicating that RL's correction can be expressed almost entirely through the output layer. We conservatively use the full rank‑32 $W_{QKVO}$ configuration for \textsc{ReasonMaxxer}.} \textit{Thus, RL's correction is not only sparse in token space but also low‑dimensional in parameter space: a tiny adapter, on the order of a fraction of a percent of the model's parameters, captures the entire distributional change.}

\vspace{-2mm}

\paragraph{From Representability to Learnability}  

The KL‑LoRA experiment shows that RL's corrective signal is representable in a tiny parameter budget. Recent work has further demonstrated that learning such a signal from scratch with LoRA‑constrained RL can match full‑parameter RL, indicating that the solution is not only low‑dimensional but also accessible within a small parameter space~\citep{wang2025tina}. This simplicity suggests that the signal might be learnable without RL's stochastic search, a hypothesis we test directly with \textsc{ReasonMaxxer} in the next section.

% =========================
% Section 5: ReasonMaxxer Method
% =========================

\section{\textsc{ReasonMaxxer} -- Entropy‑Gated Contrastive Fine‑Tuning} \label{sec:method}

\textsc{ReasonMaxxer} translates the findings of Sections~\ref{sec:post} and~\ref{sec:compress} into a direct, RL‑free training procedure.  The method generates a small set of base‑model rollouts, selects token positions where the base model's entropy is high, and applies a contrastive loss that encourages tokens leading to correct answers while penalizing those that lead to incorrect ones. The following subsections describe the problem selection, entropy‑based identification of decision points, and contrastive fine‑tuning.

\subsection{Problem Selection: Exploiting the Edge of Competence}
\label{sec:method_data}

For a collection of math problems with verifiable answers, we sample $K$ completions per problem from the frozen base model at nonzero temperature and compare each completion against the ground‑truth answer.  From this pool we keep exclusively problems where the base model’s pass rate lies strictly between~0 and~1: some rollouts are correct, others are incorrect.

This filter is the direct operationalisation of a property that both prior theoretical work~\citep{davis2025objective, zhang2025interplay} and our own oracle experiments (Section~\ref{sec:oracle}) have shown to be necessary for learning from outcome feedback.  When the base model always succeeds on a problem, there is no incorrect behaviour to penalise; when it always fails, there is no correct behaviour to reinforce.  Only the mixed‑success regime supplies the two‑sided contrastive signal that can distinguish good decisions from bad ones at the same decision points.  The filter guarantees that every retained problem contributes this signal.  In Section~\ref{sec:ablations} we verify empirically that the exact width of the pass‑rate window is not critical; the existence of both correct and incorrect rollouts within a problem is what matters.

\subsection{Decision‑Point Identification via Entropy}
\label{sec:method_entropy}

For each retained rollout we compute the per‑token entropy of the frozen base model (Eq.~\ref{eq:entropy}).  A token position $t$ is designated as a \emph{decision point} if $H_t > \tau$, where $\tau$ is a model‑family‑specific threshold chosen so that the marked positions correspond to roughly the top few percent of the model’s entropy distribution.  We write $\mathcal{D} = \{t : H_t > \tau\}$.

This step rests directly on two findings from Section~\ref{sec:post}.  First, the positions where an RL‑trained teacher disagrees with the base model are precisely the high‑entropy positions (Table~\ref{tab:sec3_taxonomy}, Fig.~\ref{fig:sec3_overview}).  Second, an entropy‑based gate can replace the teacher’s disagreement signal without loss of corrective power (Section~\ref{sec:entropy_proxy}).  Consequently, $\mathcal{D}$ is a fully teacher‑free, principled selection of the locations where the model’s behaviour most needs refinement.  Because entropy is computed from the base model alone, this stage requires no external supervision beyond the rollouts already generated.

\subsection{Advantage‑Weighted Contrastive Loss with Base Anchoring}
\label{sec:method_loss}

Given a set of rollouts for a single problem, we compute a per‑rollout normalised advantage
\begin{equation}
A_i = \frac{r_i - \bar{r}}{\sigma_r + \epsilon},
\label{eq:adv}
\end{equation}
where $r_i\!\in\!\{0,1\}$ indicates whether rollout~$i$ arrived at the correct answer, and $\bar{r},\sigma_r$ are the mean and standard deviation of the correctness indicators for that problem.  This normalisation centres the advantages so that correct and incorrect rollouts receive symmetric positive and negative weights, preventing class imbalance from distorting the gradient.

The training loss is the sum of two terms.  At decision points we apply an advantage‑weighted cross‑entropy,
\begin{equation}
\mathcal{L}_{\text{dec}} = -\sum_{t \in \mathcal{D}} A_i \cdot \log p_\theta(x_t \mid x_{<t}),
\label{eq:contrastive}
\end{equation}
which increases the likelihood of the observed token when the rollout was correct ($A_i\!>\!0$) and decreases it when the rollout was incorrect ($A_i\!<\!0$).  The model is therefore shaped to reproduce the token‑level choices that preceded a correct final answer and to avoid those that preceded an incorrect one.

At all positions \emph{outside} the decision set $\mathcal{D}$, we minimise the Kullback–Leibler divergence to the frozen base model,
\begin{equation}
\mathcal{L}_{\text{anchor}} = \sum_{t \notin \mathcal{D}} \text{KL}\big(p_{\text{base}}(\cdot \mid x_{<t}) \;\|\; p_\theta(\cdot \mid x_{<t})\big).
\label{eq:anchor}
\end{equation}
This anchor term preserves the base model’s behaviour everywhere that the mechanistic analysis found RL to have no effect, and it prevents the small adapter from overfitting to spurious correlations in the limited training set.  The total loss is $\mathcal{L} = \mathcal{L}_{\text{dec}} + \lambda \mathcal{L}_{\text{anchor}}$, the $\lambda$ balancing the two objectives.

Architecturally, \textsc{ReasonMaxxer} implements this loss through a \textbf{LoRA adapter}~\citep{hu2021lora} attached to the base model.  The base model remains frozen; only the low‑rank adapter matrices are updated.  This choice is the natural consequence of the low‑dimensionality established in Section~\ref{sec:compress}: if a rank‑32 adapter containing well under one percent of the model’s parameters can absorb RL’s entire distributional change, then the same parameter budget is more than sufficient to learn the contrastive signal directly from the base model’s own rollouts.  Implementation details are in Appendix~\ref{app:implementation}.

% =========================
% Section 6: Experiments
% =========================

\section{Experiments}
\label{sec:experiments}

We benchmark \textsc{ReasonMaxxer} on six mathematical reasoning benchmarks against publicly available RL‑trained models spanning three model families and multiple RL algorithms, and we analyze its performance, efficiency, and critical design choices

\subsection{Experimental Setup}
\label{sec:exp_setup}

\paragraph{Benchmarks and evaluation protocol.} We evaluate on six standard mathematical reasoning benchmarks: MATH‑500~\citep{hendrycks2021math}, GSM8K~\citep{cobbe2021gsm8k}, AMC~2023, AIME~2024, Minerva Math~\citep{lewkowycz2022minerva}, and OlympiadBench~\citep{he2024olympiadbench}. For AMC~2023 and AIME~2024, which contain few problems, we report \emph{avg@8} (average pass@1 over eight independent generations) to reduce variance; for all other benchmarks we report standard pass@1 from a single generation. Experiment details are given in Appendix~\ref{app:prompting}.

\paragraph{Training configuration for \textsc{ReasonMaxxer}.} We implement \textsc{ReasonMaxxer} as a rank‑32 LoRA adapter on all attention projections, leaving the base model frozen.  From a pool of 150 math problems balanced across difficulty levels we sample 20 rollouts per problem and retain 50 problems on which the base model exhibits mixed success, yielding 1000~training sequences.  Entropy‑gated decision points are selected by sweeping the threshold $\tau$ on a small held‑out set, and the adapter is trained with the advantage‑weighted contrastive loss described in Section~\ref{sec:method}.  We train for a single epoch and select the final checkpoint on a fixed 50‑problem validation split.  Full hyper‑parameters and optimizer settings are provided in Appendix~\ref{app:implementation}.

\paragraph{Cost estimation.}
To quantify efficiency, we report the estimated monetary cost of training for every method in Table~\ref{tab:main}.
For \textsc{ReasonMaxxer}, costs are directly measured from our runs on NVIDIA RTX Pro 6000 (96\,GB) GPUs using RunPod on‑demand pricing and include rollout generation, entropy scoring, the $\tau$ sweep, and checkpoint selection. Baseline costs are either taken from published reports or inferred from the official training scripts, hardware configuration, and on‑demand pricing of the corresponding GPU type (detailed in Appendix~\ref{app:cost}).
All costs are rounded to the nearest US dollar; italicised figures indicate estimates where exact numbers were not publicly documented.

\definecolor{rmcream}{RGB}{220, 232, 245}

\begin{table*}[t]
\centering
\caption{\textbf{ReasonMaxxer matches or exceeds RL baselines at dramatically lower training cost}.
For each base model, we compare against publicly released RL-trained checkpoints starting from the
same base. Best score per row in \textbf{bold}; \colorbox{rmcream}{shaded rows} report our method. Cost figures use RunPod
on-demand pricing as of submission date. Estimated costs (\textit{italicized}) are derived from each method’s
published training script and hyperparameters; see Appendix~\ref{app:cost} for full derivations.}
\label{tab:main}
\fontsize{8.8}{10}\selectfont\setlength{\tabcolsep}{4pt}
\renewcommand{\arraystretch}{1.1}
\begin{tabular}{@{}lcccccccr@{}}
\toprule
\textbf{Model} & \textbf{GSM8K} & \textbf{MATH} & \textbf{Minerva} & \textbf{Olymp.} & \textbf{AIME24} & \textbf{AMC23} & \textbf{Avg.} & \textbf{Cost} \\
& & \textbf{500} & & \textbf{Bench} & avg@8 & avg@8 & & (USD) \\
\midrule
\multicolumn{9}{c}{\textit{Qwen2.5 Family Models}} \\
\midrule
Qwen2.5-1.5B                              & 0.561         & 0.298         & 0.044         & 0.093         & 0.004         & 0.094         & 0.182         &  \\
\rowcolor{gray!10}
$\hookrightarrow$ + SimpleRL-Zoo          & \textbf{0.733} & 0.496         & 0.107         & \textbf{0.189} & \textbf{0.038} & 0.156         & \textbf{0.287} & \textit{\$200} \\
\rowcolor{gray!10}
$\hookrightarrow$ + Open-Reasoner-Zero    & 0.629         & 0.436         & 0.088         & 0.160         & 0.025         & 0.138         & 0.246         & \textit{\$1{,}200} \\
\rowcolor{rmcream}
$\hookrightarrow$ + ReasonMaxxer          & 0.710         & \textbf{0.502} & \textbf{0.114} & 0.167         & 0.029         & \textbf{0.169} & 0.282         & \textbf{\$4} \\[2pt]
Qwen2.5-7B                                & 0.751         & 0.586         & 0.129         & 0.294         & 0.046         & 0.300         & 0.351         &  \\
\rowcolor{gray!10}
$\hookrightarrow$ + SimpleRL-Zoo          & 0.851         & 0.656         & 0.132         & 0.358         & 0.088         & 0.375         & 0.410         & \$600 \\
\rowcolor{gray!10}
$\hookrightarrow$ + Open-Reasoner-Zero    & \textbf{0.924} & \textbf{0.732} & \textbf{0.213} & \textbf{0.454} & \textbf{0.121} & 0.463         & \textbf{0.485} & \textit{\$6{,}300} \\
\rowcolor{rmcream}
$\hookrightarrow$ + ReasonMaxxer          & 0.918         & 0.706         & 0.206         & 0.411         & 0.113         & \textbf{0.475} & 0.472         & \textbf{\$5} \\[2pt]
Qwen2.5-Math-7B                           & 0.450         & 0.374         & 0.088         & 0.091         & 0.092         & 0.278         & 0.229         &  \\
\rowcolor{gray!10}
$\hookrightarrow$ + SimpleRL-Zoo          & \textbf{0.827} & \textbf{0.706} & 0.151         & 0.344         & 0.188         & \textbf{0.550} & 0.461         & \$600 \\
\rowcolor{gray!10}
$\hookrightarrow$ + PRIME-Zero            & 0.543         & 0.644         & 0.132         & 0.341         & 0.171         & 0.459         & 0.382         & \textit{\$190} \\
\rowcolor{rmcream}
$\hookrightarrow$ + ReasonMaxxer          & 0.816         & 0.674         & \textbf{0.246} & \textbf{0.365} & \textbf{0.192} & 0.484         & \textbf{0.463} & \textbf{\$5} \\[2pt]
Qwen2.5-32B                               & 0.836         & 0.548         & 0.143         & 0.275         & 0.017         & 0.300         & 0.353         &  \\
\rowcolor{gray!10}
$\hookrightarrow$ + SimpleRL-Zoo          & 0.876         & 0.642         & 0.180         & 0.327         & 0.067         & 0.363         & 0.409         & \$5{,}737 \\
\rowcolor{gray!10}
$\hookrightarrow$ + Open-Reasoner-Zero    & \textbf{0.945} & \textbf{0.646} & \textbf{0.272} & 0.337         & 0.067         & 0.350         & 0.436         & \textit{\$103{,}000} \\
\rowcolor{rmcream}
$\hookrightarrow$ + ReasonMaxxer          & 0.865         & 0.636         & 0.228         & \textbf{0.356} & \textbf{0.117} & \textbf{0.438} & \textbf{0.440} & \textbf{\$25} \\
\midrule
\multicolumn{9}{c}{\textit{Mistral, DeepSeek, and Qwen3 Models}} \\
\midrule
Mistral-7B-v0.1                           & 0.070         & 0.199         & 0.056         & 0.000 & 0.007         & 0.005         & 0.056         &  \\
\rowcolor{gray!10}
$\hookrightarrow$ + SimpleRL-Zoo          & \textbf{0.076} & 0.212         & 0.025         & 0.000 & 0.011         & 0.005         & 0.055         & \$600 \\
\rowcolor{rmcream}
$\hookrightarrow$ + ReasonMaxxer          & 0.072         & \textbf{0.317} & \textbf{0.069} & 0.000 & \textbf{0.066} & \textbf{0.010} & \textbf{0.089} & \textbf{\$8} \\[2pt]
DeepSeek-R1-Distill-1.5B             & 0.775         & 0.436         & 0.136         & 0.208         & 0.092         & \textbf{0.266} & 0.319         &  \\
\rowcolor{gray!10}
$\hookrightarrow$ + DeepScaleR            & 0.793         & 0.502         & 0.162         & 0.251         & 0.075         & 0.259         & 0.340         & \$4{,}500 \\
\rowcolor{gray!10}
$\hookrightarrow$ + STILL-3               & 0.800         & 0.536         & 0.140         & 0.294         & \textbf{0.129} & 0.238         & 0.356         & \textit{\$2{,}268} \\
\rowcolor{gray!10}
$\hookrightarrow$ + Open-RS3              & 0.770         & 0.454         & 0.114         & 0.201         & 0.079         & 0.263         & 0.314         & \$42 \\
\rowcolor{rmcream}
$\hookrightarrow$ + ReasonMaxxer          & \textbf{0.825} & \textbf{0.662} & \textbf{0.213} & \textbf{0.356} & 0.117         & 0.231         & \textbf{0.401} & \textbf{\$4} \\[2pt]
Qwen3-0.6B                                & 0.489         & 0.338         & 0.059         & 0.126         & 0.000         & 0.125         & 0.189         &  \\
\rowcolor{gray!10}
$\hookrightarrow$ + GRPO       & 0.503         & 0.372         & 0.059         & 0.146         & 0.008         & 0.109         & 0.200         & \textit{\$100} \\
\rowcolor{rmcream}
$\hookrightarrow$ + ReasonMaxxer          & \textbf{0.656} & \textbf{0.470} & \textbf{0.096} & \textbf{0.179} & \textbf{0.017} & \textbf{0.253} & \textbf{0.278} & \textbf{\$4} \\[2pt]
Qwen3-4B                                  & 0.497         & 0.514         & 0.085         & 0.300         & 0.083         & 0.216         & 0.282         &  \\
\rowcolor{gray!10}
$\hookrightarrow$ + General-Reasoner      & 0.916         & \textbf{0.746} & 0.213         & \textbf{0.491} & 0.021         & 0.050         & 0.406         & \textit{\$4{,}600} \\
\rowcolor{rmcream}
$\hookrightarrow$ + ReasonMaxxer          & \textbf{0.919} & 0.660         & \textbf{0.305} & 0.403         & \textbf{0.096} & \textbf{0.472} & \textbf{0.476} & \textbf{\$4} \\
\bottomrule
\end{tabular}
\end{table*}

\subsection{Results and Analysis}
\label{sec:main_results}

Table~\ref{tab:main} presents the full set of results.
We structure the discussion around three key findings.

\paragraph{\textsc{ReasonMaxxer} matches full RL on clean comparisons.}
The most direct test compares \textsc{ReasonMaxxer} against publicly available RL trained models that were trained from the same raw base models without additional distillation or SFT stages.
On Qwen2.5‑1.5B, \textsc{ReasonMaxxer} achieves 50.2\% on MATH‑500 versus SimpleRL‑Zoo’s 49.6\% and Open‑Reasoner‑Zero’s 43.6\%, while costing \$4 compared to \$200 and \$1200 respectively.
On Qwen2.5‑7B, it reaches 70.6\% vs.\ 65.6\% (SimpleRL‑Zoo) and 73.2\% (Open‑Reasoner‑Zero), again at a tiny fraction of the cost.
The pattern holds across the Qwen2.5‑Math‑7B and Qwen2.5‑32B variants, as well as for the Mistral‑7B and Qwen3 models.
In every case, we perform on par with or better than the RL baseline while reducing training cost by two to three orders of magnitude.
This confirms that the sparse policy‑selection signal identified in Sections~\ref{sec:post}--\ref{sec:compress} is not merely a diagnostic artifact; it is \emph{the} signal that RL itself ultimately captures, and \textsc{ReasonMaxxer} recovers it without the RL optimization.

\paragraph{Performance generalizes beyond pure RL settings.}
Several baselines in Table~\ref{tab:main} incorporate additional training strategies beyond outcome‑only RL: DeepSeek‑R1‑Distill‑1.5B baselines start from a distilled checkpoint, Qwen3‑4B’s General‑Reasoner uses a model‑based verifier and multi‑domain data, and STILL‑3 employs iterative RL on a curated dataset.
Despite having access to none of these enhancements, \textsc{ReasonMaxxer} still matches or exceeds their accuracy on the majority of benchmarks. These results indicate that a substantial fraction of the gains attributed to sophisticated post‑training pipelines actually originates from the same sparse policy‑selection mechanism that \textsc{ReasonMaxxer} isolates and directly optimizes.

\paragraph{Efficiency and scalability.}
The computational and data efficiency of \textsc{ReasonMaxxer} are equally notable.
In terms of compute, \textsc{ReasonMaxxer} completes in single‑digit GPU‑hours across all models, while the RL baselines require hundreds to tens of thousands of GPU‑hours (Table~\ref{tab:main}); the average training cost is less than \$10, compared with \$100 to \$100{,}000 for the RL baselines.
In terms of data, \textsc{ReasonMaxxer} trains on 50 problems.
SimpleRL‑Zoo uses approximately 8{,}000 MATH problems for its GRPO runs, and Open‑Reasoner‑Zero trains on 57{,}000 math and reasoning problems.
This gap of over two orders of magnitude in training data is not an incidental optimisation; it follows directly from the mechanistic picture established in Sections~\ref{sec:post} and~\ref{sec:compress}.
Because RL's useful signal is concentrated at a sparse set of high‑entropy decision points, a handful of mixed‑success problems supplies sufficient contrastive supervision to capture the full policy‑steering correction.
For the same reason, \textsc{ReasonMaxxer} generates rollouts once and offline, and trains only a lightweight adapter; its cost scales with the number of training sequences rather than with the product of model size and on‑policy iterations, keeping it practical even for larger models. In summary, \textsc{ReasonMaxxer} demonstrates that the essential reasoning improvement from outcome‑based RL can be obtained by a simple, data‑efficient contrastive procedure, challenging the necessity of heavy RL infrastructure.

\subsection{Ablation Studies}
\label{sec:ablations}

\begin{figure}[ht]
\centering
\includegraphics[width=0.6\columnwidth]{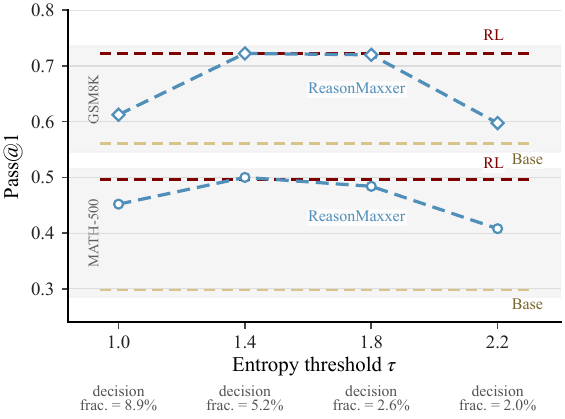}
\caption{\textbf{Sensitivity of ReasonMaxxer to the entropy threshold $\tau$.}
Pass@1 on MATH‑500 and GSM8K for Qwen2.5‑1.5B as $\tau$ varies from 1.0 to 2.2.
The percentages below the x‑axis show the mean fraction of tokens gated as decision
points at each $\tau$.  Performance is robust over a wide range: the optimal score
matches the RL model at $\tau=1.4$, and a second peak near $\tau=1.8$ aligns with
RL's observed intervention rate of 2.1\%, indicating that ReasonMaxxer does not
require a precise replication of RL's sparsity.}
\label{fig:tau_sweep}
\end{figure}

\paragraph{Sensitivity to $\tau$.}
Figure~\ref{fig:tau_sweep} shows pass@1 on MATH‑500 and GSM8K as $\tau$ varies from 1.0 to 2.2, alongside the corresponding mean fraction of tokens that are gated as decision points (annotated below each $\tau$).
Performance is robust over a broad range: the optimal MATH‑500 score (0.50) is achieved at $\tau=1.4$ (5.2\% of tokens), and the optimal GSM8K score (0.72) at the same threshold, matching the RL model.
A second peak appears at $\tau=1.8$ (2.6\% of tokens) where the decision fraction closely matches RL's observed intervention rate of 2.11\% (Table~\ref{tab:sec3_taxonomy}).
This indicates that \textsc{ReasonMaxxer} does not require a precise replication of RL's sparsity; as long as the threshold selects a plausible set of high‑entropy positions, the contrastive signal is effective.
The broad plateau confirms that entropy gating is a reliable, teacher‑free proxy for locating RL's intervention sites, as argued in Section~\ref{sec:entropy_proxy}.

\paragraph{Necessity of the contrastive term.}
To isolate the contribution of the negative gradient, we compare the full \textsc{ReasonMaxxer} loss against a variant where only correct rollouts ($A_i>0$) are used (positive‑only training, equivalent to supervised fine‑tuning on correct trajectories).
On Qwen2.5‑1.5B, positive‑only training raises MATH‑500 pass@1 from 0.298 (base) to 0.398, a non‑trivial improvement that confirms the value of targeting decision points.  However, it remains far below the RL model (0.496) and full \textsc{ReasonMaxxer} (0.502). The contrastive term that suppresses incorrect decisions thus contributes roughly half of the total gain over the base model, and it is the combination of positive reinforcement and negative suppression that together capture RL’s full policy‑steering effect.
This directly supports the design choice: a two‑sided contrastive loss exploits the edge‑of‑competence signal discussed in Section~\ref{sec:method_data}, teaching the adapter not only which tokens to prefer but also which tokens to avoid.
\section{Related Work}
\label{app:related}

Many of the works most directly relevant to this paper were already discussed in the introduction (Section~\ref{sec:intro}).  Here we provide a more complete discussion, situating our study within the broader literature.

\paragraph{What RL does for reasoning.}
A growing body of work has questioned whether RLVR expands or merely refines the base model's reasoning capabilities.
\citet{yue2025does} apply pass@$k$ analysis to show that RL‑trained models' reasoning paths lie within the base model's sampling distribution.
\citet{davis2025objective} prove that popular RL algorithms with binary rewards reduce to stochastic gradient ascent on monotone transforms of the probability of a correct answer, implying that optimisation is profitable only when the base model already succeeds non‑trivially.
\citet{zhang2025interplay} confirm this through controlled experiments, finding that RL produces genuine gains only at the model's edge of competence.
\citet{wang2025one} demonstrate that a single training example can yield large improvements, suggesting that the corrective signal RL imparts is highly compressible.
Our work provides a token‑level mechanistic characterisation that unifies these observations.

\paragraph{Entropy and decision points in LLM reasoning.}
\citet{wang2025beyond} identify high‑entropy ``forking tokens'' as the locus of RL's gradient signal and show that restricting GRPO updates to these positions matches training on all tokens.
\citet{agarwal2025entropy} demonstrate that entropy minimisation without labeled data improves reasoning performance.
\citet{park2025thinking} find that RL operates through a small number of emergent attention heads.
Together, these studies indicate that RL's effect on large language models is concentrated in a small number of structural and representational units.
The present work builds on these insights by establishing causality through oracle intervention and by showing that the sparse signal can be captured without RL.

\paragraph{RL post‑training baselines.}
Our main experiments compare against a diverse set of publicly available RL‑trained models that span multiple algorithms and training strategies.
SimpleRL‑Zoo~\citep{zeng2025simplerl} provides GRPO‑trained checkpoints across ten base models, enabling systematic comparison.
Open‑Reasoner‑Zero~\citep{hu2025openreasonerzero} scales PPO on base models without distillation, demonstrating that vanilla PPO with GAE suffices for reasoning improvement.
PRIME~\citep{cui2025prime} introduces implicit process rewards for online RL, combining outcome supervision with dense token‑level feedback.
General‑Reasoner~\citep{ma2025generalreasoner} extends GRPO with a model‑based verifier across diverse domains beyond mathematics.
On the distilled‑model track, DeepScaleR~\citep{luo2025deepscaler} applies iterative GRPO with context‑length scaling, STILL‑3~\citep{min2024still} employs a three‑stage pipeline combining imitation, exploration, and self‑improvement, and Open‑RS3~\citep{dang2025openrs3} investigates GRPO under tight compute constraints.

\paragraph{RL‑free alternatives for reasoning.}
Several methods improve reasoning without RL.
STaR~\citep{zelikman2022star} and rejection sampling fine‑tuning~\citep{yuan2023rft} train on the model's own correct solutions using uniform token‑level losses.
Best‑of‑$N$ and MCMC sampling~\citep{karan2025sampling} improve reasoning at inference time without modifying the policy.
DPO~\citep{rafailov2023dpo} offers an offline preference‑based alternative that operates at the sequence level.
The method proposed in this paper, ReasonMaxxer, differs from these approaches by explicitly targeting the sparse, entropy‑localised decision points identified in the mechanistic analysis.

\paragraph{Efficiency in reasoning.}
Making LLM reasoning more computationally efficient has been approached from several complementary angles.
On the training side, parameter‑efficient fine‑tuning methods such as LoRA~\citep{hu2021lora} and QLoRA~\citep{dettmers2023qlora} have dramatically reduced the cost of adapting large models, and a growing body of work extends these ideas to reasoning specifically: Resa~\citep{wang2025resa} uses sparse autoencoder tuning to extract reasoning abilities from a source model and guide lightweight supervised fine‑tuning, while TINA~\citep{wang2025tina} shows that LoRA‑constrained RL can match full‑parameter RL on reasoning benchmarks at a fraction of the cost.
On the inference side, a separate line of work targets the overthinking phenomenon in large reasoning models by terminating generation once sufficient confidence is reached; DEER~\citep{yang2025deer} proposes a training‑free early‑exit mechanism that monitors reasoning transition points and self‑truncates chain‑of‑thought, and LYNX~\citep{akgul2025lynx} extends this idea with lightweight hidden‑state probes and conformal prediction for distribution‑free confidence control.
Broader surveys such as~\citep{sui2025stop} provide structured taxonomies of these and related approaches.
Together, these works point toward a trend where strong reasoning performance is sought with substantially lower computational overhead than that of current RL‑heavy pipelines.
\section{Conclusion}
\label{sec:conclusion}

We set out to answer whether the RL optimization loop is necessary for improving reasoning in LLMs.  
Through systematic token‑level analysis across multiple model families and RL algorithms, we showed that RL’s useful effect on math reasoning is a sparse, predictable, and low‑dimensional correction. We then demonstrated that this correction can be obtained without RL at all with 
\textsc{ReasonMaxxer}. These results reframe reasoning improvement as a \emph{sparse policy‑selection problem}: the model already knows the necessary reasoning paths; it only needs to commit to the right branch at a handful of critical moments.  
The RL optimization loop, while capable of discovering this correction, is not a prerequisite for it.  
Our findings suggest that the community’s default investment in heavy RL infrastructure for post‑training may be disproportionate to the complexity of the problem that is actually being solved.
Recognizing this simplicity opens the door to a generation of far more efficient post‑training methods.\color{black}
\clearpage

\bibliographystyle{unsrtnat}
\bibliography{ref}

\newpage
\appendix
% =========================
% APPENDIX (drop-in)
% Place after \bibliography in main document
% =========================

\clearpage
\appendix

\section{Detailed KL‑LoRA Compression Ablations}
\label{app:kl_details}

Section~\ref{sec:compress} demonstrated that a rank‑32 QKVO adapter trained via KL distillation captures RL's full correction on reasoning tasks.
Table~\ref{tab:app_kl_ablation} reports a more aggressive compression study on Qwen2.5‑1.5B, varying both the rank and the targeted attention modules.

\definecolor{creamlight}{RGB}{255,250,240}
\begin{table}[h]
\centering
\caption{\textbf{Adapter rank and module ablation.} Even a rank‑8 adapter applied only to the output projection ($\mathbf{W}_O$) approaches the full rank‑32 QKVO adapter, indicating that RL's correction is concentrated in the output layer.}
\label{tab:app_kl_ablation}
\small
\setlength{\tabcolsep}{4pt}
\begin{tabular}{@{}lcc@{}}
\toprule
\textbf{Configuration} & \textbf{Parameters} & \textbf{Pass@1} \\
\midrule
Base model & -- & 0.233 \\
Qwen2.5‑1.5B $\rightarrow$ \textcolor[HTML]{990000}{GRPO} & 1.54B (100\%) & 0.492 \\
\midrule
\rowcolor{creamlight}
KL‑LoRA, $\mathbf{W}_{QKVO}$, rank 32 & 8.7M (0.49\%) & 0.495 \\
\rowcolor{creamlight}
KL‑LoRA, $\mathbf{W}_{QKVO}$, rank 16 & 4.4M (0.25\%) & 0.492 \\
\rowcolor{creamlight}
KL‑LoRA, $\mathbf{W}_{QKVO}$, rank 8  & 2.2M (0.12\%) & 0.487 \\
\rowcolor{creamlight}
KL‑LoRA, $\mathbf{W}_O$ only, rank 8 & 688K (0.04\%) & 0.482 \\
\bottomrule
\end{tabular}
\end{table}

The rank‑8 $\mathbf{W}_{QKVO}$ adapter already matches the RL teacher, and even an aggressively small output‑projection adapter ($\mathbf{W}_O$ only, rank 8, 688\,K parameters) lags by only 1 point on MATH‑500.
This suggests that RL's correction can be expressed almost entirely through the output layer: the base model already attends to mostly the right evidence, and RL mainly changes how the attended information is written into the hidden state to produce better next‑token choices.
We leave the exploration of such extremely compressed adapters for future work, noting that while they are sufficient to \emph{represent} RL's signal, learning the signal from scratch in such a constrained space may require different optimization strategies.
Throughout the main paper we conservatively use the full rank‑32 $\mathbf{W}_{QKVO}$ configuration for \textsc{ReasonMaxxer}.

\section{Implementation Details}
\label{app:implementation}

We provide the full training and architectural details for the KL‑LoRA distillation experiments reported in Section~\ref{sec:compress} and for \textsc{ReasonMaxxer} (Section~\ref{sec:method}).

\subsection*{KL‑LoRA Distillation}

The KL‑LoRA adapters in Section~\ref{sec:compress} are trained with a manually implemented KL divergence over the teacher’s top‑64 logits.  The teacher model (SimpleRL‑Zoo GRPO checkpoint) generates rollouts with temperature~0.6 and top‑$p$~0.95 (seed~44), and the distribution is cached.  The student (base model + LoRA) is trained for three epochs with batch size~2 and gradient accumulation~8, using the AdamW optimizer, learning rate $10^{-4}$, weight decay $10^{-2}$, and a 10\,\% warmup ratio.  The objective is averaged over generated‑token positions only.  LoRA settings for the full‑rank and compressed variants are given in Table~\ref{tab:kl_lora_config}.

\begin{table}[h]
\centering
\caption{KL‑LoRA distillation hyper‑parameters.}
\label{tab:kl_lora_config}
\small
\setlength{\tabcolsep}{3.5pt}
\rowcolors{2}{}{creamlight}
\begin{tabular}{@{}l l@{}}
\toprule
\rowcolor{creamlight}
\textbf{Hyper‑parameter} & \textbf{Value} \\
\midrule
Teacher checkpoint & Qwen2.5‑1.5B‑SimpleRL‑Zoo \\
Teacher top‑$k$ (cached) & 64 \\
Rollout temperature / top‑$p$ & 0.6 / 0.95 \\
Rollout seed & 44 \\
Epochs & 3 \\
Batch size / gradient accumulation & 2 / 8 \\
Learning rate & $10^{-4}$ \\
Weight decay & 0.01 \\
Warmup ratio & 10\,\% \\
LoRA rank (full $\mathbf{W}_{QKVO}$) & 32, 16, 8 \\
LoRA alpha (full $\mathbf{W}_{QKVO}$) & 64, 32, 16 \\
LoRA rank ($\mathbf{W}_O$‑only) & 8 \\
LoRA alpha ($\mathbf{W}_O$‑only) & 16 \\
LoRA dropout & 0.05 \\
\bottomrule
\end{tabular}
\end{table}

\subsection*{\textsc{ReasonMaxxer}}

\textsc{ReasonMaxxer} adapters are trained with AdamW using default betas $(0.9, 0.999)$ and $\epsilon=10^{-8}$, with a linear warmup followed by linear decay to zero.  All models are trained for one epoch with batch size~1 and gradient accumulation over 8 steps; the exact number of optimizer steps therefore depends on the number of training sequences, but the hyperparameters are identical across model scales.  The decision loss $\mathcal{L}_{\text{dec}}$ and the KL anchor $\mathcal{L}_{\text{anchor}}$ are each averaged over their respective token masks (decision points for $\mathcal{L}_{\text{dec}}$, all other valid prediction tokens for $\mathcal{L}_{\text{anchor}}$).  Decision points are defined exclusively on generated completion tokens; prompt tokens and padding positions are excluded.  Advantages are clipped per rollout to $[-2.5, 2.5]$ before token weighting.  The KL anchor is computed over all valid non‑decision prediction tokens in the truncated sequence (including both prompt and completion tokens), and the total loss is $\mathcal{L} = \mathcal{L}_{\text{dec}} + 0.2\,\mathcal{L}_{\text{anchor}}$.  Table~\ref{tab:reasonmaxxer_config} summarises the common hyper‑parameters.

\begin{table}[h]
\centering
\caption{Common \textsc{ReasonMaxxer} hyper‑parameters.}
\label{tab:reasonmaxxer_config}
\small
\setlength{\tabcolsep}{3.5pt}
\rowcolors{2}{}{creamlight}
\begin{tabular}{@{}l l@{}}
\toprule
\rowcolor{creamlight}
\textbf{Hyper‑parameter} & \textbf{Value} \\
\midrule
Optimizer & AdamW \\
Learning rate schedule & Linear warmup $+$ linear decay \\
Warmup steps (Qwen/DeepSeek) & 50 \\
Warmup steps (Mistral) & 30 \\
Epochs & 1 \\
Batch size / gradient accumulation & 1 / 8 \\
Gradient clipping & 1.0 \\
Weight decay (Qwen, DeepSeek) & 0.0 \\
Weight decay (Mistral) & 0.01 \\
LoRA rank / alpha / dropout & 32 / 64 / 0.0 \\
LoRA target modules & $Q, K, V, O$ \\
Max sequence length & 8192 (right truncation) \\
Rollout generation temperature / top‑$p$ & 0.6 / 0.95 \\
Advantage clipping range & $[-2.5, 2.5]$ \\
KL anchor weight $\lambda$ & 0.2 \\
Validation split & 50 problems \\
Random seed & 42 \\
\bottomrule
\end{tabular}
\end{table}

Family‑specific overrides prompt styles are listed in Table~\ref{tab:reasonmaxxer_family}. See Appendix~\ref{app:prompting} for exact templates).

\begin{table}[h]
\centering
\caption{\textsc{ReasonMaxxer} family‑specific prompt styles.}
\label{tab:reasonmaxxer_family}
\small
\setlength{\tabcolsep}{4pt}
\rowcolors{2}{}{creamlight}
\begin{tabular}{@{}l l@{}}
\toprule
\rowcolor{creamlight}
\textbf{Model} & \textbf{Prompt style} \\
\midrule
Qwen2.5‑1.5B, 7B, 32B & qwen\_boxed \\
Qwen3‑0.6B, 4B & qwen\_boxed \\
DeepSeek‑R1‑Distill‑1.5B & chat\_template \\
Mistral‑7B‑v0.1 & llama\_abel \\
\bottomrule
\end{tabular}
\end{table}

\section{Prompting and Answer Extraction}
\label{app:prompting}

The exact prompt templates and answer extraction rules used for each model family are reported below.  Within a family, the same template and extraction are applied to the base model, the RL baselines, and the \textsc{ReasonMaxxer} adapter, ensuring a fair comparison.

\subsection*{Prompt Templates}

\paragraph{Qwen2.5 and Qwen3.}
Both families use a raw completion prompt without a chat template.

\noindent\fcolorbox{gray!30}{rmcream}{%
\begin{minipage}{\dimexpr\linewidth-2\fboxsep-2\fboxrule\relax}
\texttt{Solve the following math problem step by step. Put your final answer in \textbackslash boxed\{\}.\\[2pt]
Problem: \{problem\}\\[2pt]
Solution:}
\end{minipage}}\\[4pt]

\paragraph{Mistral‑7B.}
The prompt follows a simple instruction‑answer format, without a chat template.

\noindent\fcolorbox{gray!30}{rmcream}{%
\begin{minipage}{\dimexpr\linewidth-2\fboxsep-2\fboxrule\relax}
\texttt{Question:\\ \{problem\}\\ Answer:\\ Let's think step by step.}
\end{minipage}}

\paragraph{DeepSeek‑R1‑Distill.}
The native chat template is applied with a single user message.

\noindent\fcolorbox{gray!30}{rmcream}{%
\begin{minipage}{\dimexpr\linewidth-2\fboxsep-2\fboxrule\relax}
\texttt{Solve the following math problem step by step. Give a concise solution and put your final answer in \textbackslash boxed\{\}.\\[2pt]
Problem: \{problem\}}
\end{minipage}}\\[4pt]

\section{Cost Estimation Details}
\label{app:cost}

Table~\ref{tab:main} reports monetary training costs for all methods.  Here we describe how those figures were obtained.  Whenever a baseline paper explicitly reports wall‑clock time and hardware, we use those numbers directly.  When such information is not published, we infer GPU‑hours from the official training scripts and documented hyperparameters, then convert to cost using RunPod on‑demand pricing as of Apr~28, 2026 (\$2.49 per H100‑hour for H100‑80G instances; \$2.69 per H100‑hour for H100‑SXM instances where specified; \$1.89 per GPU‑hour for RTX Pro 6000 instances).  In all cases we note whether the cost figure is a direct measurement or an estimate.

\paragraph{ReasonMaxxer.}
All ReasonMaxxer costs are direct measurements on the authors' hardware (4$\times$ NVIDIA RTX Pro 6000 Blackwell, 96\,GB GDDR7).  The reported GPU‑hours include rollout generation, entropy scoring, a sweep over $\tau \in \{1.2, 1.4, 1.6, 1.8\}$, training, and final checkpoint selection on a hold‑out set.  Cost is computed at the RunPod on‑demand rate for that GPU type.

\paragraph{SimpleRL‑Zoo.}
For Qwen2.5‑7B and larger models, SimpleRL‑Zoo~\citep{zeng2025simplerl} directly reports the number of GPUs and training hours.  For the 1.5B variant, which does not have a separately published wall‑clock, we estimate GPU‑hours by scaling the per‑step generation time of the 7B run by model size, holding the training configuration constant (1024~prompts $\times$ 8~rollouts $\times$ $\sim$100 GRPO steps)\footnote{The public training command is available at \url{https://github.com/hkust-nlp/simpleRL-reason}.}.  The Mistral‑7B run uses the same hardware and step count as the Qwen2.5‑7B group.

\paragraph{Open‑Reasoner‑Zero.}
Open‑Reasoner‑Zero~\citep{hu2025openreasonerzero} does not report wall‑clock for any model size.  We estimate GPU‑hours from the public PPO recipes in the project repository\footnote{See \texttt{playground/orz\_1p5b\_ppo.py} and \texttt{playground/orz\_7b\_ppo.py} in the official Open‑Reasoner‑Zero codebase: \url{https://github.com/Open-Reasoner-Zero/Open-Reasoner-Zero}.}.  The estimation procedure uses the documented hardware configuration (number of nodes, GPUs per node), the number of prompts and rollouts per step, and the step counts inferred from the training curves in the paper.  Per‑step time is calibrated against SimpleRL‑Zoo's published figures for a comparable model size, with a 1.5--2$\times$ overhead factor to account for the PPO critic and GAE computation absent in GRPO.  The final point estimates are reported as midpoints of plausible ranges; italicised costs in Table~\ref{tab:main} reflect this estimation.

\paragraph{PRIME‑Zero.}
PRIME‑Zero~\citep{cui2025prime} publishes per‑step wall‑clock time and the total number of RL steps.  We multiply the two to obtain GPU‑hours and convert using the RunPod rate for the reported GPU type.  The training data size uses the number of prompts consumed during RL, approximately 13K out of the full Eurus‑2‑RL dataset.

\paragraph{General‑Reasoner (Qwen3‑4B).}
Ma et al.~\citep{ma2025generalreasoner} state that the 4B model is trained on 4 nodes $\times$ 8 H100 GPUs for around 2~days, giving $\sim$1,536 H100‑hours.  We adopt this figure directly.

\paragraph{DeepSeek‑based baselines.}
DeepScaleR~\citep{luo2025deepscaler} self‑reports 3,800 A100‑hours and a total cost of $\sim$\$4,500, which we use as given.  STILL‑3~\citep{min2024still} does not provide hardware details; the cost figure in Table~\ref{tab:main} is taken from the comparison table in Open‑RS3~\citep{dang2025openrs3}, which estimates 1,200 A100‑hours on 8$\times$ A100‑80GB.  Open‑RS3 self‑reports 96 A40‑hours, which we convert to cost at the RunPod A40 rate.

\paragraph{Mistral‑7B.}
The SimpleRL‑Zoo run for Mistral‑7B uses the same hardware and step count as the Qwen2.5‑7B group, with simplified prompts.  We therefore assign it the same GPU‑hour estimate (240 H100‑hours).  The training data is 8K Easy‑split problems.

In all cases, the cost figures in Table~\ref{tab:main} are rounded to the nearest US dollar.  The italicisation indicates estimates where the exact wall‑clock was not directly published by the original authors.

\end{document}